\documentclass[10pt, twocolumn, letterpaper]{article}

\usepackage[pagenumbers]{res/cvpr}      

\usepackage{float}
\usepackage{bbding}
\usepackage{pifont}
\usepackage{graphicx}
\graphicspath{{./res/figs/}}
\usepackage{amsmath}
\usepackage{amssymb}
\usepackage{comment}
\usepackage{booktabs}
\usepackage{multirow}
\usepackage{makecell}
\usepackage{titletoc}
\usepackage{setspace}
\usepackage{appendix}
\usepackage[table]{xcolor}
\usepackage[hang,flushmargin]{footmisc}

\usepackage[pagebackref, breaklinks, colorlinks]{hyperref}

\usepackage[capitalize]{cleveref}
\crefname{section}{Sec.}{Secs.}
\Crefname{section}{Section}{Sections}
\Crefname{table}{Table}{Tables}
\crefname{table}{Tab.}{Tabs.}

\makeatletter
\newcommand\figcaption{\def\@captype{figure}\caption}
\newcommand\tabcaption{\def\@captype{table}\caption}
\makeatother

\newcommand{\hiddensection}[1]{
    \refstepcounter{section}
    \section*{\arabic{section}.\hspace{1ex}{#1}}
}
\newcommand{\hiddensubsection}[1]{
    \refstepcounter{subsection}
    \subsection*{\arabic{section}.\arabic{subsection}.\hspace{1ex}{#1}}
}
\newcommand{\hiddensubsubsection}[1]{
    \refstepcounter{subsubsection}
    \subsubsection*{\arabic{section}.\arabic{subsection}.\arabic{subsubsection}.\hspace{1ex}{#1}}
}

\begin{document}

\title{OvarNet: Towards Open-vocabulary Object Attribute Recognition}
\author{Keyan Chen$^{1\star}$, Xiaolong Jiang$^{2\star}$, Yao Hu$^2$, Xu Tang$^2$, Yan Gao$^2$, Jianqi Chen$^1$, Weidi Xie$^{3\dag }$ \\[3pt]
Beihang University$^1$, \hspace{5pt} Xiaohongshu Inc$^2$, \hspace{5pt} Shanghai Jiao Tong University$^3$\\[2pt]
\url{https://kyanchen.github.io/OvarNet}
}

\twocolumn[{
\maketitle
\vspace{-1cm}
\begin{figure}[H]
\hsize=\textwidth
\centering
\includegraphics[width=\textwidth]{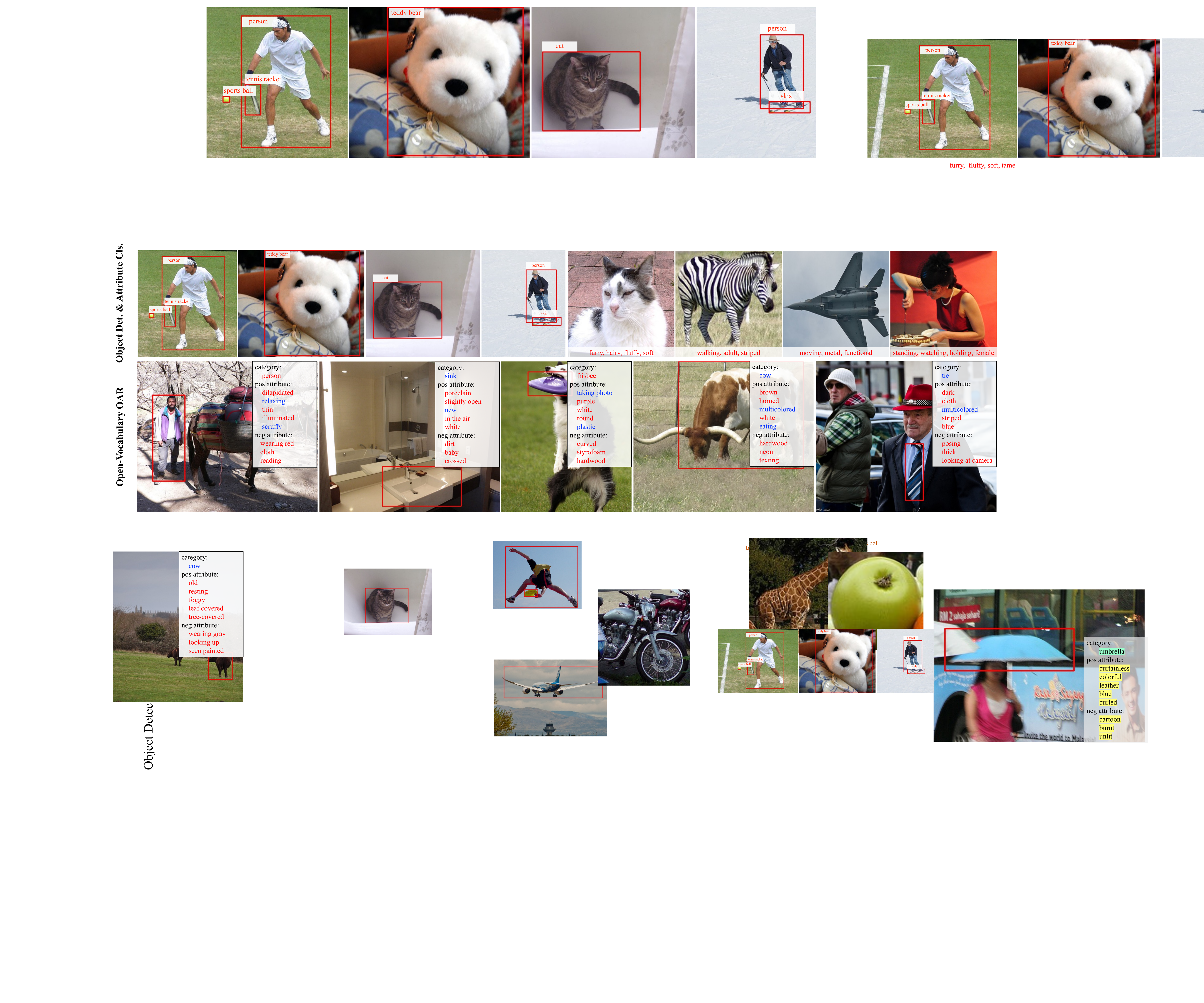}
\vspace{-0.7cm}
\caption{The first row depicts the tasks of object detection and attribute classification in a close-set setting, \textit{i.e.}, train and test on the same vocabulary set. 
The second row gives qualitative results from our proposed OvarNet, 
which simultaneously localizes, categorizes, and characterizes arbitrary objects in an open-vocabulary scenario. We only show one object per image for ease of visualization, \textcolor[rgb]{1,0,0}{\textbf{red}} denotes the base category/attribute \textit{i.e.}, 
seen in the training set, while \textcolor[rgb]{0.07,0.19,0.96}{\textbf{blue}} represents the novel category/attribute unseen in the training set.
}
\label{fig:ovar}
\end{figure}
}]

\let\thefootnote\relax\footnotetext{$\star$ Equal contribution. $\dag$ Corresponding author.}

\begin{abstract}
In this paper, we consider the problem of simultaneously detecting objects and inferring their visual attributes in an image, even for those with no manual annotations provided at the training stage, resembling an open-vocabulary scenario.
To achieve this goal, we make the following contributions: 
(i) we start with a na\"ive two-stage approach for open-vocabulary object detection and attribute classification, termed CLIP-Attr. The candidate objects are first proposed with an offline RPN and later classified for semantic category and attributes;
(ii) we combine all available datasets and train with a federated strategy to finetune the CLIP model, aligning the visual representation with attributes,
additionally, we investigate the efficacy of leveraging freely available online image-caption pairs under weakly supervised learning;
(iii) in pursuit of efficiency, we train a Faster-RCNN type model end-to-end with knowledge distillation, that performs class-agnostic object proposals and classification on semantic categories and attributes with classifiers generated from a text encoder;
Finally, (iv) we conduct extensive experiments on VAW, MS-COCO, LSA, and OVAD datasets, 
and show that recognition of semantic category and attributes is complementary for visual scene understanding, {\em i.e.}, jointly training object detection and attributes prediction largely outperform existing approaches that treat the two tasks independently, 
demonstrating strong generalization ability to novel attributes and categories. 
\end{abstract}

\hiddensection{Introduction}\label{sec:intro}

Understanding the visual scene in terms of objects has been the main driving force for development in computer vision, for example, in object detection, the goal is to localise objects in an image and assign one of the pre-defined semantic labels to them, such as a `car’, `person’ or `bus’, despite tremendous success has been made by the community, such task definition has largely over-simplified our understanding of the visual world, as a visual object can often be characterised from many aspects other than semantic category, for example, a bus can be `yellow' or `black', a shirt can be `striped' or `unpatterned', learning attributes can thus complement category-level recognition, acquiring more comprehensive visual perception.

In the literature, numerous work has shown that understanding the objects' attributes can greatly facilitate object recognition and detection, 
even with few or no examples of visual objects~\cite{farhadi2009describing, lampert2009learning, Meng_2020_CVPR, zhu2020attribute, xu2020attribute},
for example, Farhadi \textit{et al.} proposed to shift the goal of object recognition from `naming' to `description', which allows naming familiar objects with attributes, but also to say something about unfamiliar objects~(``hairy and four-legged'', not just ``unknown'')  \cite{farhadi2009describing}; Lampert \textit{et al.} considered the open-set object recognition, 
that aims to recognise objects by human-specified high-level description, 
{\em e.g.}, arbitrary semantic attributes, like shape, color, or even geographic information, instead of training images \cite{lampert2009learning}.
However, the problem considered in these seminal work tends to be a simplification from today's standard, for example, attribute classification are often trained and evaluated on object-centric images under the close-set scenario, 
{\em i.e.}, assuming the bounding boxes/segmentation masks are given~\cite{pham2021learning, isola2015discovering, saini2022disentangling}, or sometimes even the object category are known as a prior~\cite{metwaly2022glidenet, pham2021learning}.

In this paper, we consider the task of simultaneously detecting objects and classifying the attributes in an open-vocabulary scenario, {\em i.e.}, the model is only trained on a set of base object categories and attributes, 
while it is required to generalise towards ones that are unseen at training time, as shown in Fig.~\ref{fig:ovar}. 
Generally speaking, we observe three major challenges:
{\em First}, in the existing foundation models, \textit{e.g.}, CLIP~\cite{radford2021learning} and ALIGN~\cite{jia2021scaling}, 
the representation learned from image-caption pairs tends to bias towards object category, rather than attributes, which makes it suffer from feature misalignment when used directly for attribute recognition. We experimentally validate this conjecture by showing a significant performance drop in attribute recognition, compared to category classification;
{\em Second}, there is no ideal training dataset with three types of annotations, object bounding boxes, semantic categories, and attributes; as far as we know, 
only the COCO Attributes dataset~\cite{patterson2016cocoattr} provides such a degree of annotations, but with a relatively limited vocabulary size (196 attributes, 29 categories);
{\em Third}, training all three tasks under a unified framework is challenging and yet remains unexplored, \textit{i.e.}, simultaneously localising~(`where’), classifying objects' semantic categories and attributes~(`what’) under the open-vocabulary scenario.

To address the aforementioned issues,
we start with a na\"ive architecture, termed as CLIP-Attr, 
which first proposes object candidates with an offline RPN~\cite{ren2015faster},
and then performs open-vocabulary object attribute recognition by comparing the similarity between the attribute word embedding and the visual embedding of the proposal. To better align the feature between attribute words and proposals, we introduce learnable prompt vectors with parent attributes on the textual encoder side and finetune the original CLIP model on a large corpus of the freely available image-caption datasets. 
To further improve the model efficiency, we present OvarNet, 
a unified framework that performs detection and attributes recognition at once, 
which is trained by leveraging datasets from both object detection and attribute prediction, as well as absorbing knowledge from CLIP-Attr to improve the performance and robustness of unseen attributes.
As a result, our proposed OvarNet, being the first scalable pipeline, can simultaneously localize objects and infer their categories with visual attributes in an open-vocabulary scenario. Experimental results demonstrate that despite only employing weakly supervised image-caption pairs for distillation, OvarNet outperforms previous the state-of-the-art on VAW~\cite{pham2021learning}, MS-COCO~\cite{lin2014microsoft}, LSA~\cite{pham2022improving} and OVAD~\cite{bravo2022open} datasets, exhibiting strong generalization ability on novel attributes and categories.

\hiddensection{Related Work}

\noindent \textbf{Attribute Prediction.}
Visual attribute aims to describe one object/scene from various aspects, 
for example, color, texture, shape, material, state, etc,
allowing to represent object categories in a combinatorial manner.
However, annotating attributes can be very time-consuming, 
early efforts only focus on specific domains such as fashion~\cite{zhang2020texture,  zhang2019task},  face~\cite{huang2019deep,  zhong2016face}, animals~\cite{abdulnabi2015multi,wah2011caltech}, 
posing severe limitations for real-world deployment. 
With the release of large-scale datasets including COCO Attributes~\cite{patterson2016cocoattr}, Visual Genome~\cite{krishna2017visual}, and VAW~\cite{pham2021learning},  recent work considers building models for large-vocabulary attributes classification ~\cite{pham2021learning, yun2022attributes}. Nonetheless, these methods only perform multi-class classification on pre-computed image patches,  which not only fail to acquire object localization ability but also endure extra computation overhead due to redundant feature extraction passes. Additionally, other methods such as SCoNE ~\cite{pham2021learning} require object category as input to perform attribute prediction, leading to extra complexity in practice. In this work, we aim to build a unified framework that can jointly settle object localization,  category prediction, and attribute prediction in an open-vocabulary scenario, relieving the aforementioned practical limitations. 

\begin{figure*}[h]
\centering
\includegraphics[width=\linewidth]{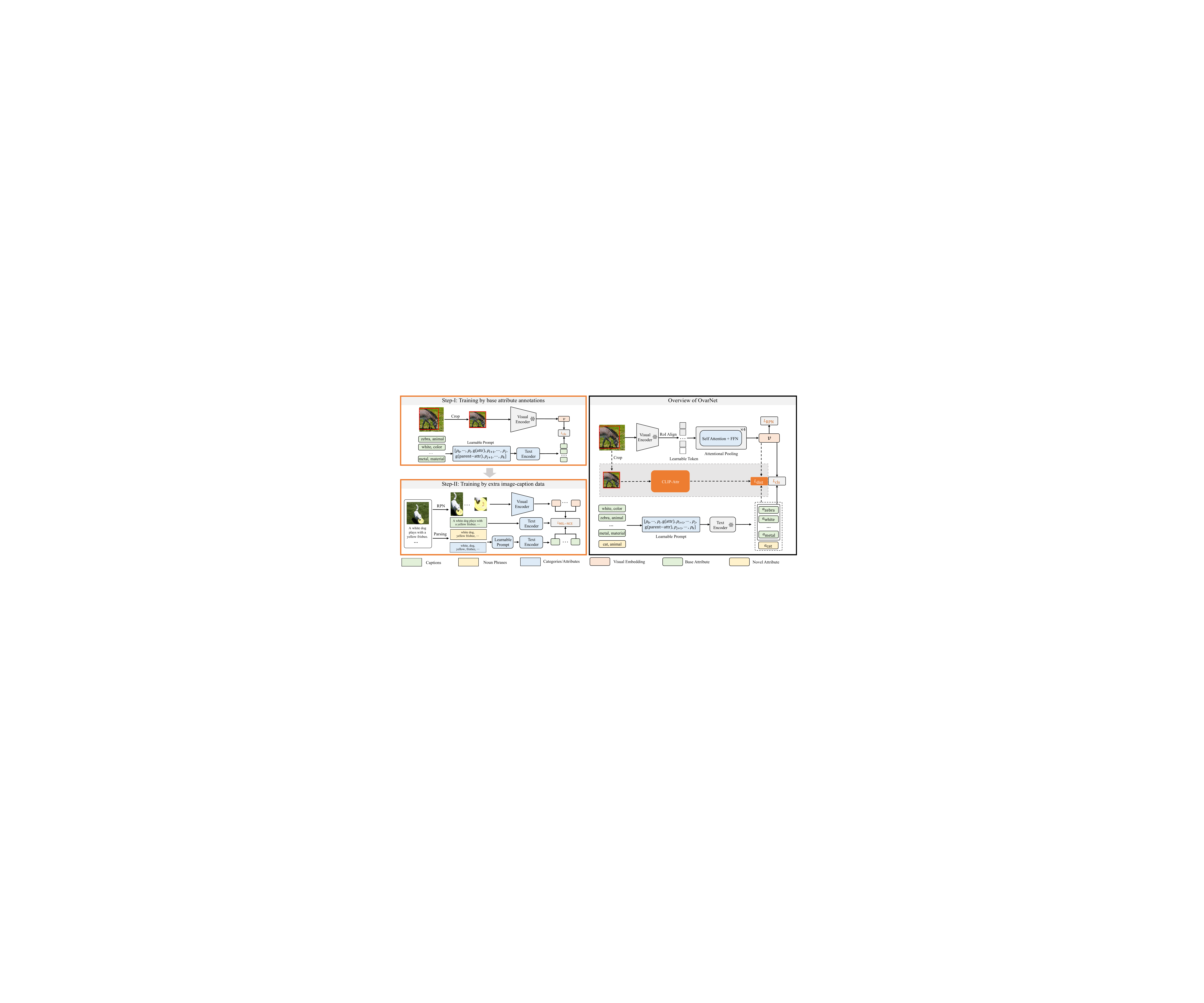}
\caption{An overview of the proposed method.
\textbf{Left:} the two-step training procedure for finetuning the pre-trained CLIP to get CLIP-Attr that better aligns the regional visual feature to attributes. \textbf{Step-I:} na\"ive federate training by base attribute annotations. \textbf{Step-II:} training by image-caption pairs. We first conduct RPN on the whole image to get box-level crops, parse the caption to get noun phrases, categories, and attributes, and then match these fine-grained concepts for weakly supervised training.
 \textbf{Right:} the proposed one-stage framework OvarNet. We inherit the CLIP-Attr for open-vocabulary object attribute recognition. Regional visual feature is learned from the attentional pooling of proposals; while attribute concept embedding is extracted from the text encoder. Solid lines declare the standard federated training regime. Dashed lines denote training by knowledge distillation with CLIP-Attr.
 }
\label{fig:ovdet}
\vspace{-10pt}
\end{figure*}

\vspace{3pt}
\noindent \textbf{Open-vocabulary Object Detection.}
Open-vocabulary object detection strives to detect all objects,  
including those that are unseen at the training stage. 
Existing approaches~\cite{feng2022promptdet,  bansal2018zero,  zhou2022detecting,  gu2021open} achieve open-vocabulary capability by replacing the detector's classifier with object category word embedding from the pre-trained visual-language model, {\em e.g.}, CLIP, and perform category classification via embedding matching. 
In specific, OVR-CNN~\cite{zareian2021open} proposes an efficient training approach with image-caption pairs that can be easily obtained from the website. 
ViLD \cite{gu2021open} adopts distillation to infuse open-vocabulary knowledge into a two-stage detector, Detic~\cite{zhou2022detecting} increases the size of detector's vocabulary to twenty-thousand by exploiting the large dataset~(ImageNet-21K) with image-level annotations. PromptDet~\cite{feng2022promptdet} leverages the pre-trained CLIP~\cite{radford2021learning} and aligns the detector's visual embedding with text embedding with learnable prompts. However, none of these models considers simultaneously inferring attributes for detected objects.

\vspace{3pt}
\noindent \textbf{Zero-shot Learning.}
Zero-shot learning aims to extend the model's capability towards recognising objects beyond those seen categories at training time~\cite{bansal2018zero,  gu2021open,  rahman2018zero,  rahman2020improved}. In the context of object detection,  
early zero-shot solutions rely on visual attributes to infer unseen categories~\cite{mao2020zero,  jayaraman2014zero,  xu2020attribute,  lampert2013attribute},  aiming to represent category by attributes, such that it can generalize from seen to unseen category. Recent methods adopt vision-language feature alignment to achieve zero-shot learning, based on similarity computation between the visual feature and text concepts.
\hiddensection{Methodology}

In this section, we start by introducing the problem scenario~(Sec.~\ref{sec:problem-scenario}), 
followed by describing a na\"ive architecture for open-vocabulary attribute classification by steering a pre-trained CLIP model, dubbed CLIP-Attr~(Sec.~\ref{sec:two-stage}), and finally, we further distill the knowledge from CLIP-Attr into a more efficient two-stage detection architecture called OvarNet, which can perform detection and attribute prediction in a unified framework~(Sec.~\ref{sec:student}).

\hiddensubsection{Problem Scenario} 
\label{sec:problem-scenario}
Assuming we are given a training dataset, 
\textit{i.e.}, $\mathcal{D}_{\text{train}} = \{(\mathcal{I}_1, y_1), \dots, (\mathcal{I}_N, y_N)\}$, where $\mathcal{I}_i \in \mathbb{R}^{H \times W \times 3}$ refers to an image, 
and $y_i = \{b_i, c_i, a_i\}$ denotes its corresponding ground-truth annotations, 
with the coordinates for $n$ object bounding boxes~($b_i \in \mathbb{R}^{n_i \times 4}$), 
their corresponding semantic categories~($c_i \in \mathbb{R}^{n_i \times \mathcal{C}_{\text{base}}}$), and a set of binary attributes for each object~($a_i \in \{0, 1\}^{n_i \times \mathcal{A}_{\text{base}}}$).
Our goal is to train a model that can process any image from a test set~($\mathcal{I}_k \sim \mathcal{D}_{\text{test}}$), simultaneously localising the objects and inferring their semantic categories, and visual attributes:
\begin{align*}
    \{\hat{b}_k, \hat{c}_k, \hat{a}_k\} = \Phi_{\text{CLS}} \circ \Phi_{\text{LOC}} (\mathcal{I}_k)
\end{align*}
where the image is progressively processed by a class-agnostic object localization, and open-vocabulary attributes classification, to produce the $\hat{b}_k \in \mathbb{R}^{n_k \times 4}, \hat{c}_k \in \mathbb{R}^{n_k \times \mathcal{C}_{\text{test}}}$ and $\hat{a}_k \in \{0,1\}^{n_k \times \mathcal{A}_{\text{test}}}$.
Note that, at inference time, 
the objects may be of unseen/novel semantic categories or attributes, \textit{i.e.}, $\mathcal{C}_{\text{test}} = \mathcal{C}_{\text{base}} \cup \mathcal{C}_{\text{novel}}$,  $\mathcal{A}_{\text{test}} = \mathcal{A}_{\text{base}} \cup \mathcal{A}_{\text{novel}}$, thus the considered problem falls into open-vocabulary object attributes recognition.
For simplicity, we will omit the subscript $k$ while describing the proposed models. 
To avoid redundancies, we treat the category as a super-attribute for modeling our pipeline unless otherwise specified.

\vspace{-5pt}
\hiddensubsection{Two-stage Object Attribute Recognition}
\label{sec:two-stage}

In this section, we describe a two-stage open-vocabulary attribute classification method, 
termed CLIP-Attr, that first uses a class-agnostic region proposal network (RPN) to generate object candidates, then verifies the candidates with category and attributes using a finetuned CLIP:
\begin{align*}
\begin{split}
\{\hat{b}_k\} &= \Phi_{\text{LOC}} =  \Phi_{\text{crpn}}(\mathcal{I}) \\
\{\hat{c}_k, \hat{a}_k\} &= \Phi_{\text{CLS}} =  \Phi_{\text{cls}} \circ \Phi_{\text{clip-v}} \circ \Phi_{\text{crop}} (\mathcal{I}, \{\hat{b}_k\})
\end{split}
\end{align*}
where $\Phi_{\text{crpn}}(\mathcal{I})$ is a class-agnostic RPN, $\Phi_{\text{cls}}(\cdot)$ represents attributes classification, $\Phi_{\text{clip-v}}(\cdot)$ denotes the CLIP visual encoder, and $\Phi_{\text{crop}}(\cdot)$ is an operation that crops $\hat{b}_k$ the box region from input image.

\vspace{-3pt}
\hiddensubsubsection{Object-centric Visual Encoding}

\noindent \textbf{Class-agnostic Region Proposal.}
\label{sec:rpn}
To propose the candidate regions that potentially have objects situated, 
we employ a Faster-RCNN \cite{ren2015faster} based region proposal network that parametrises the anchor classification and bounding box regression in a class-agnostic manner, 
\textit{i.e.}, $\Phi_{\text{crpn}}(\cdot)$ shares parameters for all categories. 
Inspired by the observation in \cite{zhou2022detecting,  feng2022promptdet, kaul2022label},
we train the proposal network only on base categories offline, and it shows sufficient generalization ability towards unseen categories.

\vspace{5pt}
\noindent \textbf{RoI Visual Pooling.}
\label{sec:teacher-visual}
Given the pre-defined object boxes, we acquired the image crops~($\Phi_{\text{crop}}(\cdot)$) and feed them into the CLIP image encoder ($\Phi_{\text{clip-v}}(\cdot)$) to compute regional visual embeddings $\hat{v}_i \in \mathbb{R}^{1 \times D}$, $i$ denotes the $i$th region.

\hiddensubsubsection{Open-vocabulary Attributes Classification}
\label{sec:teacher-cls}

\vspace{2pt}
\noindent \textbf{Generating Attribute Embedding.}
\label{sec:teacher_text}
To compute attribute embeddings, we employ the pre-trained text encoder from CLIP~($\Phi_{\text{clip-t}}(\cdot)$), and use two variants of prompts for better aligning the attribute with the visual region features:
(i) for each attribute, we employ prior knowledge of ontologies, and encode their parent-class words along with the attribute, for example, the embedding for the `wet' attribute can be expanded as:  $\Phi_{\text{clip-t}} (\text{wet}, \text{state})$ to better distinguish from  $\Phi_{\text{clip-t}} (\text{water}, \text{material})$, 
or $\Phi_{\text{clip-t}}(\text{in water}, \text{place})$;
(ii) we augment it with multiple learnable prompt vectors,
as a consequence, the attribute embeddings can be computed as:
\begin{align}
\begin{split}
    \hat{t}_j &= \Phi_{\text{clip-t}}([p_0,  \cdots,  p_i,  \text{g}(\text{attribute}), p_{i+1},  \cdots, p_j,\\
    & \text{g}(\text{parent-attribute}),  p_{j+1},  \cdots, p_k])
\end{split}
\end{align}
where $\text{g}(\cdot)$ denotes the tokenisation procedure, 
and $p_i~(i \in {0,  1,  \cdots,  k})$ has the same dimension with the attribute word embeddings, denoting the learnable prompt vectors, that are shared across all attributes, and can generalize towards unseen attributes at inference time.

\vspace{4pt}
\noindent \textbf{Attribute Classification.} 
\label{sec:teacher_cls}
Attribute prediction can be obtained by computing the similarity between visual region feature and attribute concept embedding as:
\begin{align}
    \hat{s}_{ij} = \Phi_{\text{cls}}(\hat{v}_i, \hat{t}_j) = 
    \sigma(\langle \hat{v}_i^T, \hat{t}_j \rangle/ \tau), \label{eq:similarity}
\end{align}
where both $v_i$ and $t_j$ are L2 normalised, 
and $\hat{s}_{ij}$ denotes the likelihood that the $i$th region contains the $j$th attribute. $\tau$ is a temperature parameter and $\sigma$ denotes sigmoid function.

\hiddensubsubsection{Training Procedure}
\label{sec:teacher_train}
In this section, we describe the training procedure for open-vocabulary attributes classification, which strives to better align the regional visual feature to the attribute description.

\vspace{4pt}
\noindent \textbf{Step-I: Federated Training.}

\label{sec:align1}
In order to align the regional visual feature to attributes, 
an ideal training dataset should contain three types of annotations, 
namely, object bounding boxes, semantic categories, and attributes,
as far as we know, the COCO Attributes dataset~\cite{patterson2016cocoattr} is the only one that provides such a level of annotations, but with a very limited vocabulary size~(196 attributes, 29 categories).

To fully exploit the annotations in existing datasets, we combine the detection dataset, 
\textit{e.g.,}~COCO \cite{lin2014microsoft}, and attribute prediction dataset, \textit{e.g.,}~VAW \cite{pham2021learning}. Specifically, we follow standard procedure for training the class-agnostic region proposal network with images from COCO, {\em i.e.}, SmoothL1 loss and Binary Cross Entropy (BCE) are applied for box coordinates regression and objectness prediction; while for training attribute/category classification, 
as illustrated in the top-left part of Fig.~\ref{fig:ovdet}, 
we employ ground-truth bounding boxes to crop the objects, and compute their visual embeddings with a pre-trained visual encoder from CLIP, we finetune CLIP's \textbf{text encoder} by optimising BCE loss with multi-label attribute classification, as follows,
\begin{align}
\begin{split}
    \mathcal{L}_{\text{cls}} &= \frac{1}{N} \sum\nolimits_{i=1}^{N}
    w_i \cdot \text{BCE}(\hat{s}_i,  s_i)
\end{split} \label{eq:teacher_cls}
\end{align}

\noindent 
where $N = |\mathcal{C}_{\text{base}}|+|\mathcal{A}_{\text{base}}|$, denotes the class number of categories and attributes, $i$ denotes the $i$-th category/attribute,  $\hat{s}_i$ is the predicted probability, and $s_i \in \{0,  1, \text{unk}\}$ denotes an attribute label being negative, positive or missing. By default, for the missing attributes we treat them as negative with a re-weight factor, \textit{i.e.}, $s_i = 0$ during training. $w_i \propto 1/{f_i}^\gamma$, $\sum_{i=1}^N w_i = N$, where $f_i$ indicates the occurrence frequency of the $i$-th attribute in the training set, $\gamma = 0.25$ is a smoothing factor.
As a result, this step ends up with a finetuned CLIP text encoder that better aligns the regional visual feature to attributes, referred to $\Phi_{\text{CLIP-Attr}}(\cdot)$.

\vspace{4pt}
\noindent \textbf{Step-II: Training with Image-caption Dataset.}
\label{sec:alignment2}
To further improve the alignment, especially for novel attributes, 
we also consider using freely available image-caption datasets, \textit{e.g.}, $\mathcal{D}_{\text{img-cap}} = \{ \{\mathcal{I}_1,s_1\}, \dots, \{\mathcal{I}_N, s_N\}\}$, 
where the $\mathcal{I}_i, s_i$ refer to image and caption sentence respectively. 
We detect all the objects in each image with a class-agnostic object proposal as described in Sec.~\ref{sec:rpn}. 
We keep the largest box proposal~($b^*$) and those with top-K objectness scores~($b^k$), 
and crop original images with the inferred bounding boxes.
We pass these crops through $\Phi_{\text{CLIP-Attr}}(\cdot)$, 
to get the predictions for semantic categories and attributes,
and keep those with confidence scores higher than 0.7 as pseudo-positive labels. 
In addition, for caption sentences, we use TextBlob~\cite{TextBlob} to parse all captions into `semantic category', `attribute', 
and `noun phrases' based on COCO and VAW dictionaries. 
For example, the sentence ``A striped zebra is eating green grass"
is processed and converted to \{category: `zebra'\}, 
\{attribute: `green', `striped'\}, \{noun phrase: `striped zebra', `green grass'\}.

To this end, we continue finetuning the alignment model~($\Phi_{\text{CLIP-Attr}}(\cdot)$) 
with the pseudo groundtruths obtained from the pre-processing stage.
In detail, we compute the visual and textual embeddings as in \textbf{Step-I}, 
however, as the labels obtained from captions or the model's prediction are not guaranteed to be correct, that requires special actions.
We adopt multi-instance contrastive learning~(MIL-NCE)~\cite{miech2020end}, 
that maximizes the accumulated similarity score of positive matches between the visual and textual embeddings as follows:
\begin{align}
\small
\mathcal{L}_{\text{MIL-NCE}} = - \log 
\frac{\sum\limits_{(v,t)\in \mathcal{P}}\text{\hspace{-3pt}exp}(\frac{\langle v^{T}\text{\hspace{-3pt}, }t \rangle}{\tau})}
{\sum\limits_{(v,t)\in \mathcal{P}}\text{\hspace{-3pt}exp}(\frac{\langle v^{T}\text{\hspace{-3pt}, }t \rangle}{\tau}) \text{\hspace{2pt}} + \text{\hspace{-10pt}}
\sum\limits_{(v^\prime,t^\prime) \sim \mathcal{N}}\text{\hspace{-5pt}exp}(\frac{\langle {v^\prime}^{T}\text{\hspace{-3pt}, } t^{\prime} \rangle}{\tau})} \label{eq:mil_loss}
\end{align}

\noindent 
where $\mathcal{P}$ is a set of \textit{positive} pairs with image crop feature and textual concept embeddings, $\mathcal{N}$ conversely refers to an associated set of \textit{negative} pairs. Here, we pair the largest box ($b^*$) with the given caption, 
{\em i.e.}, noun phrases, attributes, and semantic categories. 
While for the other top-K boxes ($b^k$), we treat the {\bf model inferred} categories and attributes as positives. 
Here, we continue training both \textbf{visual and text encoders} in $\Phi_{\text{CLIP-Attr}}$ by optimising the following loss:
\begin{align}
\small
\mathcal{L}_{\text{cls}} = 1/K \cdot \sum_{k=0}^{K} \mathcal{L}_{\text{MIL-NCE}}^k
\label{eq:teacher_alignment2_loss}
\end{align}
where $\mathcal{L}_{\text{MIL-NCE}}^k$ denotes MIL-NCE loss over the $k$th box and the corresponding textual concepts (here, we treat the largest box $b^*$ as the $0$th). 
An overview is shown in the bottom-left of Fig. \ref{fig:ovdet}.

\hiddensubsection{Distilled Object Attribute Recognition}
\label{sec:student}

Although open-vocabulary object attribute prediction can be realised by the above proposed $\Phi_{\text{CLIP-Attr}}$ with the pre-computed proposals, the inference procedure is time-consuming, because every cropped region is fed into the visual encoder. 
In this section, we aim to address the slow inference speed, 
and train a Faster-RCNN type model end-to-end for object detection and attribute prediction,
termed as OvarNet~(Open-vocabulary attribute recognition):
\begin{align*}
    \{\hat{b}_k, \hat{c}_k, \hat{a}_k\} = \Phi_{\text{Ovar}} = \Phi_{\text{cls}} \circ \Phi_{\text{crpn}} \circ \Phi_{\text{v-enc}}(\mathcal{I})
\end{align*}

\noindent 
where the image is sequentially processed by a visual encoder, class-agnostic region proposal, and open-vocabulary attributes classification, as illustrated in the right of Fig. \ref{fig:ovdet}.

\vspace{5pt}
\noindent {\bf Visual Encoder.}
To start with, the input image is fed into a visual backbone, 
obtaining multi-scale feature maps:
\begin{align}
    \mathcal{F} = \{f^1, \dots, f^l\} = \Phi_{\text{v-enc}}(\mathcal{I}) 
\end{align}
where $f^i$ refers to the feature map at $i$-th level, we adopt the visual encoder from $\Phi_{\text{CLIP-Attr}}$.

\vspace{5pt}
\noindent {\bf Class-agnostic Region Encoding.}
\label{sec:student_rpn}
To extract regional visual embeddings for candidate objects, 
we make the class-agnostic region encoding as follows,
\begin{align}
    \{\hat{v}_1, \dots, \hat{v}_n\} = \Phi_{\text{crpn}} = \Phi_{\text{attn-pool}} \circ \Phi_{\text{roi-align}} \circ \Phi_{\text{rpn}}(\mathcal{F})
\end{align}
specifically, the feature pyramid is used in the region proposal network to fuse multi-scale features. The ROI-align's output~($\mathbb{R}^{14 \times 14 \times 256}$) is firstly down-sampled with a convolutional layer~(stride $2$ and kernel size $2 \times 2$), and then passed into a block with 4 Transformer encoder layers with a learnable token, 
acting as attentional pooling. 
As a result, $\hat{v}_i \in \mathbb{R}^{1 \times D}$ 
refers to the feature embedding of the $i$-th candidate object. 
We train the proposal network only on base categories as described in Sec.~\ref{sec:rpn}

\vspace{5pt}
\noindent {\bf Open-vocabulary Attributes Classification.}
\label{sec:student-attr-cls}
We extracted attribute concept embeddings as in Sec.~\ref{sec:teacher_text}. 
After obtaining the embeddings for each of the proposed objects, 
we can classify them into arbitrary attributes or categories by measuring the similarity between visual and attribute embeddings~(Eq.~\ref{eq:similarity}).

\vspace{5pt}
\noindent {\bf Federated Training.}
\label{sec:student_train}
We combine both COCO and VAW, and adopt a similar federated training strategy as in CLIP-Attr, 
with the key difference being that we jointly supervise localization for class-agnostic region proposal and classification for attribute prediction. 
The overall loss function can be formulated as: $\mathcal{L}_{\text{total}} = \mathcal{L}_{\text{cls}}  + \lambda_{\text{RPN}} \cdot \mathcal{L}_{\text{RPN}}$, $\lambda_{\text{RPN}}$ is a re-weighted parameter.

Intuitively, if the embedding spaces for visual and textual can be well-aligned by training on a limited number of base categories/attributes, the model should enable open-vocabulary object attribute recognition with the aforementioned training procedure, 
however, in practice, we observe unsatisfactory performance on the novel categories and attributes. We further incorporate additional knowledge distillation from the CLIP-Attr model described in Sec.~\ref{sec:teacher_train} to improve the model's ability for handling unseen categories and attributes.

\vspace{5pt}
\noindent {\bf Training via Knowledge Distillation.}
\label{sec:distillation}
In addition to the federated training loss $\mathcal{L}_{\text{total}}$, we introduce an extra distillation item $\mathcal{L}_{\text{dist}}$, that encourages similar prediction 
between $\Phi_{\text{CLIP-Attr}}(\cdot)$ and $\Phi_{\text{Ovar}}(\cdot)$:
\begin{align}
\mathcal{L}_{\text{dist}}(\hat{s}, s) = \frac{1}{N} \sum\nolimits_{i=1}^{N}\text{KL}(\hat{s}_{i}, s_{i}), 
\end{align}
where $\hat{s}$ is prediction probabilities over all attributes from OvarNet and $s$ is the prediction by using image crops from the aligned $\Phi_{\text{CLIP-Attr}}$. $\text{KL}$ denotes the Kullback-Leibler divergence loss.

\hiddensection{Experimental Setup}
\hiddensubsection{Datasets}
\label{sec:data}
Here, we introduce the datasets for training and evaluation of our proposed models for open-vocabulary object attribute recognition. Note that, while training the model, 
we have to consider two aspects of the openness evaluation, one is on semantic category, 
and the other is on the attributes.

\vspace{4pt}
\noindent \textbf{MS-COCO}~\cite{lin2014microsoft}.
We follow the setup for generalized zero-shot detection as proposed in ZSD~\cite{bansal2018zero}: 48 classes are selected as base classes ($\mathcal{C}_{\text{base}}$), and 17 classes are used as unseen/novel classes ($\mathcal{C}_{\text{novel}}$). 
The train and minival sets are the same as standard MS-COCO 2017. 
At the training stage, only the images with base category objects are used. 

\vspace{3pt}
\noindent \textbf{VAW}~\cite{pham2021learning}.
For attributes recognition, VAW is constructed with VGPhraseCut~\cite{wu2020phrasecut} and GQA~\cite{hudson2019gqa}, containing a large vocabulary of 620 attributes, for example, color, material, shape, size, texture, action, {\em etc}. Each instance is annotated with positive, negative, and missing attributes. In our experiments, we sample half of the `tail' attributes and 15\% of the `medium' attributes as the novel set ($\mathcal{A}_{\text{novel}}$, 79 attributes) and the remaining as the base ($\mathcal{A}_{\text{base}}$, 541 attributes). 
More details are included in the supplementary material.

\vspace{4pt}
\noindent \textbf{Image-Caption Datasets}.
Conceptual Captions 3M (CC-3M)~\cite{sharma2018conceptual} contains 3 million image-text pairs harvested from the web with wide diversities, and COCO Caption~(COCO-Cap)~\cite{chen2015microsoft} comprises roughly 120k images and 5-way image-caption curated style annotations. We only keep images whose pairing captions have overlapped attributes or categories in the COCO and VAW dictionaries. 
We refer to the two subsets as CC-3M-sub and COCO-Cap-sub.

\vspace{4pt}
\noindent \textbf{LSA~\cite{pham2022improving}}.
A recent work by Pham \textit{et al.} proposed the Large-Scale object Attribute dataset~(LSA). 
LSA is constructed with all the images and their parsed objects and attributes of the Visual Genome (VG)~\cite{krishna2017visual}, GQA~\cite{hudson2019gqa}, COCO-Attributes~\cite{patterson2016cocoattr}, Flickr30K-Entities~\cite{plummer2015flickr30k}, MS-COCO~\cite{lin2014microsoft}, and a portion of Localized Narratives (LNar)~\cite{pont2020connecting}. Here, we evaluate the effectiveness of our proposed method with the same settings proposed in the original paper: LSA common (4921 common attributes for the base set, 605 common attributes for the novel set); LSA common $\rightarrow$ rare (5526 common attributes for the base set, 4012 rare attributes for the novel set).

\vspace{4pt}
\noindent \textbf{OVAD~\cite{bravo2022open}}.
OVAD introduces the open-vocabulary attributes detection task with a clean and densely annotated attribute evaluation benchmark (no training set is provided). 
The benchmark defines 117 attribute classes for over 14,300 object instances.

\vspace{4pt} 
\noindent \textbf{Summary}.
We have constructed the COCO-base and VAW-base datasets for training, and COCO-novel and VAW-novel for evaluation purposes, with the former for {\em object category classification}, 
and the latter for {\em object attributes classification}.
To align regional visual features with attributes in CLIP-Attr, 
we use COCO-base and VAW-base for \textbf{Step-I} training, 
and then use CC-3M-sub and COCO-Cap-sub for \textbf{Step-II} finetuning. 
Later, COCO-base and VAW-base are employed in distilling knowledge from CLIP-Attr to OvarNet for efficiency. On the OVAD benchmark, training data is not provided, we directly evaluate the OvarNet that is trained with COCO, VAW, and COCO-Cap-sub. On the LSA dataset, we train OvarNet with the base attribute annotations in LSA common and LSA common $\rightarrow$ rare for evaluation purposes. We refer the reader to a more detailed table with dataset statistics in the supplementary material.

\begin{table*}[t]
\begin{minipage}[t]{0.43\linewidth}
\centering
\small
\vspace{0pt}
\resizebox{\textwidth}{!}{
  \begin{tabular}{*{3}{c} | *{2}{c} | *{2}{c}}
  \toprule
    \multicolumn{1}{c}{\multirow{2}{*}{\textbf{Attribute}}} & \multicolumn{1}{c}{\multirow{1}{*}{\textbf{Parent}}} & \multicolumn{1}{c|}{\multirow{2}{*}{\textbf{M/L}}} & \multicolumn{2}{c|}{\multirow{1}{*}{\textbf{VAW}}} & \multicolumn{2}{c}{\multirow{1}{*}{\textbf{COCO}}}\\
    \multicolumn{1}{c}{} & \multicolumn{1}{c}{\textbf{Attribute}} & \multicolumn{1}{c|}{} & $\textbf{AP}_\textbf{novel}$ & $\textbf{AP}_\textbf{all}$ & $\textbf{AP}_\textbf{novel}$ & $\textbf{AP}_\textbf{all}$\\
    \midrule
    \checkmark & \ding{55} & none & 52.15 & 59.16 & 40.53 & 49.84\\
    \checkmark & \ding{55} & M & 53.64 & 62.22 & 41.65 & 52.35\\
    \checkmark & \checkmark & M & 53.78 & 62.76 & 41.97 & 52.81\\
    \checkmark & \ding{55} & L & 55.73 & 64.54 & 42.77 & 53.80\\
    \checkmark & \checkmark & L & \cellcolor{gray!30} 57.39 & \cellcolor{gray!30} 66.92 & \cellcolor{gray!30} 45.82 & \cellcolor{gray!30} 55.21\\
    \bottomrule
  \end{tabular}
  }
  \vspace{-5pt}
  \caption{\small{Ablation study on prompt engineering with CLIP-Attr model. 
  \textbf{M/L} denotes whether manually designed prompts or learnable prompts are used.}}
\label{tab:prompt}
\end{minipage}
\hspace{8pt}
\begin{minipage}[t]{0.57\linewidth}
\centering
\small
\vspace{0pt}
\resizebox{\textwidth}{!}{
  \begin{tabular}{c   c | *{3}{c} | *{3}{c}}
  \toprule
    \multicolumn{1}{c}{\multirow{2}{*}{\textbf{Method}}}  &\multicolumn{1}{c|}{\multirow{2}{*}{\textbf{Training Data}}} & \multicolumn{3}{c|}{\multirow{1}{*}{\textbf{VAW}}} & \multicolumn{3}{c}{\multirow{1}{*}{\textbf{COCO}}}\\
    \multicolumn{1}{c}{} & \multicolumn{1}{c|}{} & $\textbf{AP}_\textbf{base}$ & $\textbf{AP}_\textbf{novel}$ & $\textbf{AP}_\textbf{all}$ &  $\textbf{AP}_\textbf{base}$ & $\textbf{AP}_\textbf{novel}$ & $\textbf{AP}_\textbf{all}$ \\
    \midrule
    Plain CLIP  &none & 47.69 & 46.15 & 47.53 & 38.56 & 41.13 & 39.45\\
    \midrule
    $\Phi_{\text{CLIP-Attr}}$ &COCO-base & 49.03 & 47.07 & 48.75 & \cellcolor{gray!30}59.33 & 42.49 & 53.93 \\
    $\Phi_{\text{CLIP-Attr}}$ & VAW-base & 67.71 & 57.28 & 66.82 & 38.90 & 42.54 & 39.98 \\
    $\Phi_{\text{CLIP-Attr}}$ & COCO-base $+$ VAW-base & \cellcolor{gray!30}67.90 & \cellcolor{gray!30}57.39 & \cellcolor{gray!30}66.92 & 58.26 & \cellcolor{gray!30}45.82 & \cellcolor{gray!30}55.21 \\
    \midrule
    $\Phi_{\text{CLIP-Attr}}$ & $+$ CC-3M-sub & 69.79 & \cellcolor{gray!30}59.16 & 68.87 & 65.79 & 48.90 & 61.36\\
    $\Phi_{\text{CLIP-Attr}}$ & $+$ COCO-Cap-sub & \cellcolor{gray!30}70.24 & 57.73 & \cellcolor{gray!30}69.03 & \cellcolor{gray!30}69.62 & \cellcolor{gray!30}52.61 & \cellcolor{gray!30}65.17 \\
    \bottomrule
  \end{tabular}
}
\vspace{-6pt}
\caption{\small{Oracle test for Step-I and Step-II training with \textbf{objects' boxes given}.
`Plain CLIP' directly classifies cropped images with a manual prompt.}}
\label{tab:alignment}
\end{minipage}
\vspace{-0.3cm}
\end{table*}

\hiddensubsection{Evaluation Protocol and Metrics}
Our considered open-vocabulary object attribute recognition involves two sub-tasks: open-vocabulary object detection and classifying the attributes for all detected objects. 
We evaluate the two sub-tasks in both \textbf{box-given} and \textbf{box-free} settings on COCO for category detection and VAW, LSA, and OVAD for attribute prediction. 
Specifically, the box-given setting is widely used in attribute prediction and object recognition communities~\cite{pham2021learning, saini2022disentangling, uijlings2013selective, girshick2014rich}, where the ground-truth bounding box annotations are assumed to be available for all objects, and the protocol only evaluates object category classification and multi-label attribute classification with mAP metric; In contrast, the box-free setting favors a more challenging problem, as the model is also required to simultaneously localise the objects, and classify the semantic category and attributes. 

Note that, the annotations on existing attribute datasets, such as VAW, LSA, 
are \textbf{not exhaustive or object-centric}, (i) not all the objects are labeled in an image, (ii) some annotations are on stuffs, that represents uncountable amorphous regions, such as sky and grass. We have to strike a balance in the box-free setting for attributes by matching the predicted boxes to the ground-truth box with the largest IoU, and then evaluate the attribute predictions using mAP. We consider the aforementioned metrics over base set classes, novel set classes, and all classes.

\begin{table}[t]
  \centering
  \small
  \resizebox{\linewidth}{!}{
  \begin{tabular}{*{4}{c} | *{2}{c} | *{2}{c}}
  \toprule
     \multicolumn{4}{c|}{\multirow{1}{*}{\textbf{MIL-NCE}}} &\multicolumn{2}{c}{\multirow{1}{*}{\textbf{VAW}}} & \multicolumn{2}{c}{\multirow{1}{*}{\textbf{COCO}}}\\
     \textbf{$b^*$-cap.} & \textbf{$b^*$-phr.} & \textbf{$b^*$-attr.} & \textbf{$b^k$-attr.} &  $\textbf{AP}_\textbf{novel}$ & $\textbf{AP}_\textbf{all}$ & $\textbf{AP}_\textbf{novel}$ & $\textbf{AP}_\textbf{all}$\\
    \midrule
     &  &  & &  57.39 & 66.92 & 45.82 & 55.21\\
    \checkmark &  &  & &  57.45 & 66.94 & 45.87 & 55.36\\
    \checkmark & \checkmark &  &  &   57.42 & 67.87 & 48.29 & 57.92\\
    \checkmark & \checkmark & \checkmark & &  57.61 & \cellcolor{gray!30}69.33 & 51.83 & 63.80 \\
    \checkmark & \checkmark & \checkmark & \checkmark &  \cellcolor{gray!30}57.73 & 69.03 & \cellcolor{gray!30}52.61 & \cellcolor{gray!30}65.17\\
    \bottomrule
  \end{tabular}
  }
  \vspace{-5pt}
  \caption{\small{The effect of different weakly supervised loss terms in Step-II training. 
  We conduct ablation studies with COCO-Cap-sub dataset. 
  \textbf{$b^*$} and \textbf{$b^k$} refer to the largest object proposal and top-K objectness proposals of an image respectively.
  } }\label{tab:alignment_loss}
  \vspace{-0.5cm}
\end{table}

\hiddensubsection{Implementation details} 

\noindent \textbf{CLIP-Attr Training.}
We use the pre-trained R50-CLIP as the visual backbone to get object-centric visual features; 
all cropped regions are resized to $224 \times 224$ based on the short side with the original aspect ratio kept. Similar to Detic~\cite{zhou2022detecting}, we use sigmoid activation and multi-label binary cross-entropy loss for classification.
We adopt the Stochastic Gradient Descent (SGD) optimizer with a learning rate of 0.001, a weight decay of 0.0001, and a momentum of 0.9. In \textbf{Step-I} training, we train the prompt vectors and text encoder for 50 epochs. In \textbf{Step-II} training with the image-caption dataset, we further finetune the entire model~(both visual and textual encoders) for another 40 epochs. We select 30 top-K proposals while pre-possessing COCO-Cap-sub, and 15 for CC-3M-sub, 
as the images are often object-centric in the latter case.

\vspace{4pt}
\noindent \textbf{OvarNet Training.}
In OvarNet, we initialise its visual backbone with the trained CLIP-Attr~(Resnet50 without AttentionPool2d layer) and keep its text encoder frozen for efficiency. We adopt the AdamW optimizer with a learning rate of 0.0001. The models are trained for 30 epochs with the distillation term, and 60 epochs without distillation. We employ 640-800 scale jittering and horizontal flipping, and the temperature parameter $\tau$ is configured to be trainable. Following the observation and related prior, we empirically set: $\gamma = 0.25$, and $\lambda_{\text{RPN}} = 1$. All the experiments are conducted on 8 NVIDIA A100 GPUs.

\vspace{4pt}
\noindent \textbf{Prompt Engineering.}
We have experimented with different numbers of prompt vectors,
empirically, we take 30 vectors and divide them into 10 inserting before, between, and after the attribute and the parent-class attributes. 
In terms of prompts used for encoding noun phrases,
we use 16 learnable vectors, {\em i.e.}, 8 before and 8 after phrase embedding.

\begin{table*}[t]
\small
\centering
\begin{minipage}[t]{0.275\linewidth}
\centering
\vspace{0pt}
\resizebox{\textwidth}{!}{
  \begin{tabular}{l | *{2}{c} | *{2}{c}}
  \toprule
    \multicolumn{1}{c|}{\multirow{2}{*}{\textbf{Distil.}}} &\multicolumn{2}{c|}{\multirow{1}{*}{\textbf{VAW}}} & \multicolumn{2}{c}{\multirow{1}{*}{\textbf{COCO}}}\\
    \multicolumn{1}{c|}{} & $\textbf{AP}_\textbf{novel}$ & $\textbf{AP}_\textbf{all}$ & $\textbf{AP}_\textbf{novel}$ & $\textbf{AP}_\textbf{all}$\\
    \midrule 
    none & 50.53 & 61.74 & 30.43 & 59.83 \\
    Feat. L2 & 51.87 & 63.34 & 33.17 & 59.16 \\
    Feat. L1 & 52.57 & 64.65 & 32.92 & 59.62 \\
    Prob. KL & \cellcolor{gray!30}56.43 & \cellcolor{gray!30}68.52 & \cellcolor{gray!30}54.10 & \cellcolor{gray!30}67.23 \\
    \bottomrule
  \end{tabular}
  }
  \vspace{-5pt}
  \caption{\small{Ablation study on knowledge distillation \textbf{with boxes given}.}} \label{tab:dist}
\end{minipage}
\hfill
\begin{minipage}[t]{0.335\linewidth}
\centering
\vspace{0pt}
\resizebox{\textwidth}{!}{
  \begin{tabular}{l c | *{2}{c} | *{2}{c}}
  \toprule
    \multicolumn{1}{c}{\multirow{2}{*}{\textbf{Model}}} & \multicolumn{1}{c|}{\multirow{2}{*}{\textbf{Arch.}}} & \multicolumn{2}{c|}{\multirow{1}{*}{\textbf{VAW}}} & \multicolumn{2}{c}{\multirow{1}{*}{\textbf{COCO}}}\\
    \multicolumn{1}{c}{} & \multicolumn{1}{c|}{} &  $\textbf{AP}_\textbf{novel}$ & $\textbf{AP}_\textbf{all}$ & $\textbf{AP}_\textbf{novel}$ & $\textbf{AP}_\textbf{all}$ \\
    \midrule
    $\Phi_{\text{CLIP-Attr}}$  & R50 & \cellcolor{gray!30}57.73 & 69.03 & 52.61 & 65.17 \\
    $\Phi_{\text{CLIP-Attr}}$  & ViT-B/16 & 57.69 & \cellcolor{gray!30}71.72 & \cellcolor{gray!30}58.30 & \cellcolor{gray!30}70.94\\
    \midrule
    $\Phi_{\text{Ovar}}$  & R50 & \cellcolor{gray!30}56.43 & 68.52 & 54.10 & 67.23 \\
    $\Phi_{\text{Ovar}}$  & ViT-B/16 & 56.41 & \cellcolor{gray!30}68.79 & \cellcolor{gray!30}55.70 & \cellcolor{gray!30}68.02 \\
    \bottomrule
  \end{tabular}
  }
  \vspace{-6pt}
  \caption{\small{Experiments with different $\Phi_{\text{CLIP-Attr}}$ architectures in a \textbf{box-given setting}.}} \label{tab:arch}
\end{minipage}
\hfill
\begin{minipage}[t]{0.355\linewidth}
\centering
\vspace{0pt}
\resizebox{\textwidth}{!}{
  \begin{tabular}{l c | *{2}{c} |  *{2}{c}}
  \toprule
    \multicolumn{1}{c}{\multirow{2}{*}{\textbf{Initialisation}}} & \multicolumn{1}{c|}{\multirow{2}{*}{\textbf{Freeze}}} & \multicolumn{2}{c|}{\multirow{1}{*}{\textbf{VAW}}} & \multicolumn{2}{c}{\multirow{1}{*}{\textbf{COCO}}}\\
    \multicolumn{1}{c}{} & \multicolumn{1}{c|}{} & $\textbf{AP}_\textbf{novel}$ & $\textbf{AP}_\textbf{all}$ & $\textbf{AP50}_\textbf{novel}$ & $\textbf{AP50}_\textbf{all}$ \\
    \midrule
    ImageNet & \ding{55} & 50.27 & 59.22 & 31.06 & 48.62\\
    $\Phi_{\text{CLIP-Attr}}$ & \ding{55} & 54.85 & \cellcolor{gray!30}68.02 & \cellcolor{gray!30}35.25 & \cellcolor{gray!30}54.31\\
    $\Phi_{\text{CLIP-Attr}}$ & \checkmark & \cellcolor{gray!30}55.47 & 67.62 & 35.17 & 54.15\\
    \bottomrule
  \end{tabular}
  }
  \vspace{-6pt}
  \caption{\small{Ablation study on finetuning the visual backbone of OvarNet with different initializations under \textbf{box-free setting}.}} \label{tab:init}
\end{minipage}
\end{table*}

\begin{figure*}
\centering
\hspace{-5pt}
\begin{minipage}[t]{0.66\linewidth}
\centering
\vspace{-94pt}
\resizebox{\textwidth}{!}{
  \begin{tabular}{l | c | *{3}{c} | *{3}{c}}
  \toprule
    \multicolumn{1}{c|}{\multirow{2}{*}{\textbf{Method}}} & \multicolumn{1}{c|}{\multirow{2}{*}{\textbf{Training Data}}} & \multicolumn{3}{c|}{\multirow{1}{*}{\textbf{VAW}}} &  \multicolumn{3}{c}{\multirow{1}{*}{\textbf{COCO}}}\\
    
    \multicolumn{1}{c|}{} & \multicolumn{1}{c|}{} & $\textbf{AP}_\textbf{base}$ & $\textbf{AP}_\textbf{novel}$ & $\textbf{AP}_\textbf{all}$ &  $\textbf{AP50}_\textbf{base}$ & $\textbf{AP50}_\textbf{novel}$ & $\textbf{AP50}_\textbf{all}$ \\
    \midrule
    SCoNE \cite{pham2021learning}& fully supervised & - & - & 68.30 & - & - & - \\
    TAP \cite{pham2022improving} & fully supervised & - & - & 65.40 & - & - & -\\
    OVR-RCNN \cite{zareian2021open}& COCO Cap & - & - & - & 46.00 & 22.80 & 39.90 \\
    OVR-RCNN \cite{zareian2021open}& CC 3M& - & - & - & - & - & 34.30 \\
    ViLD \cite{gu2021open}& CLIP400M & - & - & - & 59.50 & 27.60 & 51.30 \\
    Region CLIP \cite{zhong2022regionclip}& COCO Cap & - & - & - & 54.80 & 26.80 & 47.50\\
    Region CLIP \cite{zhong2022regionclip}& CC 3M & - & - & - & 57.10 & 31.40 & 50.40\\
    PromptDet \cite{feng2022promptdet} & Web Images & - & - & - & - & 26.60 & 50.60\\
    Detic \cite{zhou2022detecting} & COCO Cap & - & - & - & 47.10 & 27.80 & 45.00\\
    \midrule
    OvarNet (box-given) & COCO-base + VAW-base & 68.27 & 53.75 & 66.85 & 60.94 & 41.44 & 55.85 \\
    OvarNet (box-given)& $+$CC 3M-sub & 69.30 & 55.44 & 67.96 & 68.35 & 52.34 & 64.18 \\
    OvarNet (box-given) & $+$COCO Cap-sub & \cellcolor{gray!30}\textbf{69.80} & \cellcolor{gray!30}\textbf{56.43} & \cellcolor{gray!30}\textbf{68.52} & \cellcolor{gray!30}\textbf{71.88} & \cellcolor{gray!30}\textbf{54.10} & \cellcolor{gray!30}\textbf{67.23} \\
    \midrule
    OvarNet (box-free)& COCO-base + VAW-base & 67.71 & 53.42 & 66.03 & 56.20 & 32.02 & 49.77\\
    OvarNet (box-free)& $+$CC 3M-sub & 67.32 & 54.26 & 66.75 & 59.50 & 33.68 & 52.40 \\
    OvarNet (box-free)& $+$COCO Cap-sub & \cellcolor{gray!30}\textbf{68.93} & \cellcolor{gray!30}\textbf{55.47} & \cellcolor{gray!30}\textbf{67.62} &  \cellcolor{gray!30}\textbf{60.35} & \cellcolor{gray!30}\textbf{35.17} & \cellcolor{gray!30}\textbf{54.15}\\ 
    \bottomrule
  \end{tabular}
  }
  \vspace{-6pt}
  \tabcaption{Comparison for open-vocabulary object detection and attribute prediction on the VAW test set and COCO validation.}\label{tab:sota}
  \end{minipage}
\hfill
\vspace{2pt}
\begin{minipage}[h]{0.33\linewidth}
\centering
\includegraphics[width=\textwidth]{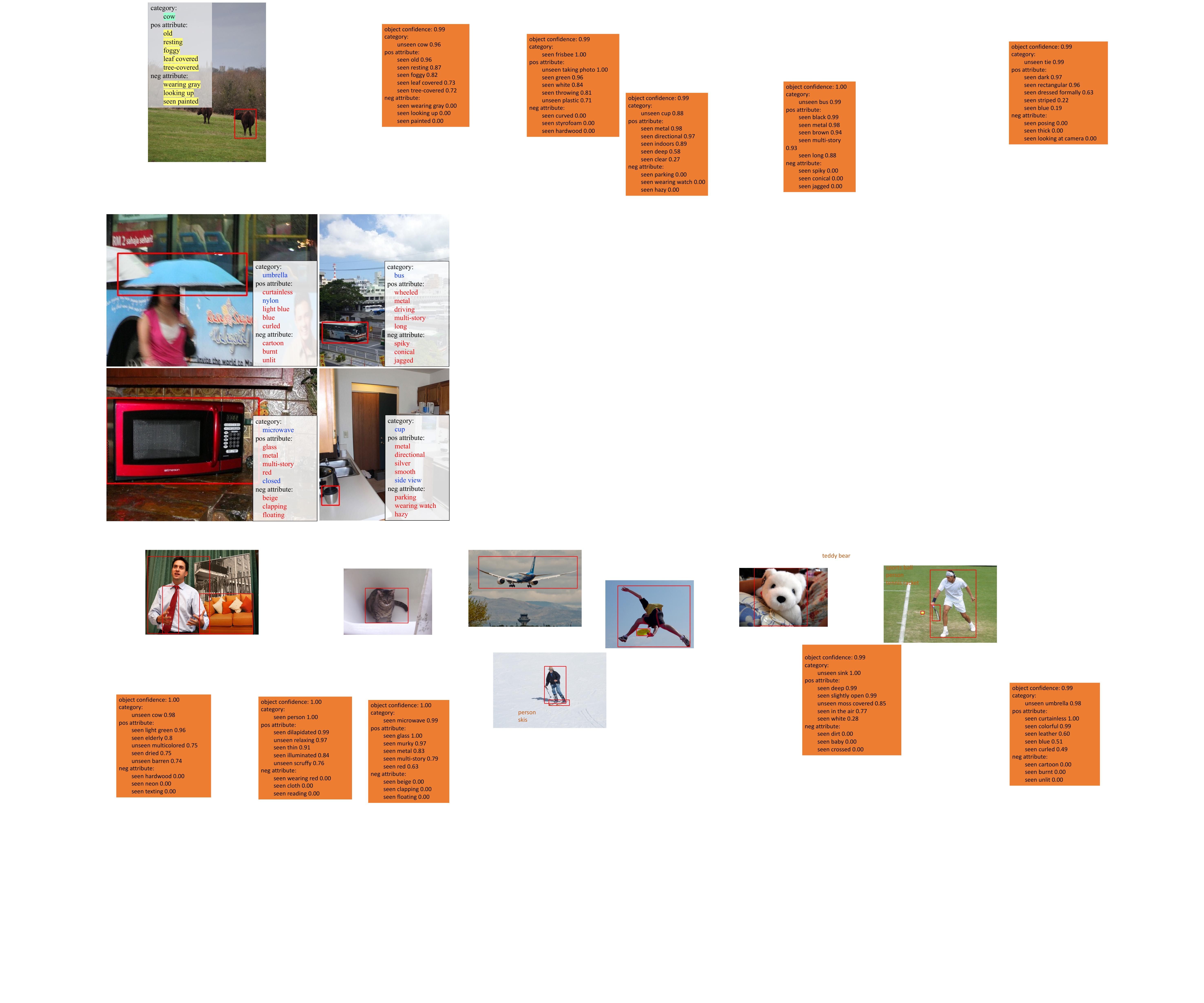}
\vspace{-18pt}
\figcaption{Qualitative visualization from OvarNet. 
\textcolor[rgb]{1,0,0}{\textbf{Red}}: base category/attributes. \textcolor[rgb]{0.07,0.19,0.96}{\textbf{Blue}}: category/attributes.}
\label{fig:results}
\end{minipage}
\vspace{-10pt}
\end{figure*}

\hiddensubsection{Ablation Study}
We conduct ablation studies on VAW and COCO datasets, 
to thoroughly validate the effectiveness of proposed components, 
including prompt learning with the parent-class attribute, different losses for training CLIP-Attr,
and the effect of \textbf{Step-I} and \textbf{Step-II} training. 
Finally, we validate the effectiveness of knowledge distillation.

\vspace{4pt}
\noindent \textbf{Prompt Learning with Parent-class Attribute.}
In attribute embedding, we employ two variants of prompts for better aligning the attribute with the visual region features. We compare the learned prompt to the manual prompt while training $\Phi_{\text{CLIP-Attr}}$ with the pre-annotated object boxes. 
As shown in Tab.~\ref{tab:prompt}, comparing to the results from only using plain attribute words, using carefully designed prompt~\cite{zhou2022learning, gu2021open}, for example, ``It is a photo of [\underline{category}]" and ``The attribute of the object is [\underline{attribute}]" for the category and attribute words, indeed delivers improvements; 
While adding the parent-class word to the prompt template, 
\textit{i.e.}, use the ``The attribute of the object is [\underline{attribute}], and it is a [\underline{parent-attribute}]”, the lexical ambiguity can be alleviated, leading to a considerable improvement on novel categories and attributes. Finally, our proposed prompt learning with parent-class attribute words further brings a performance improvement by 3.61/3.85 AP on novel attributes/categories, compared to the manual prompt with parent-class words.

\vspace{4pt}
\noindent \textbf{Effect of Step-I and Step-II Training.}
We first compare the performance of attributes classification on regional visual feature~(assuming ground-truth object boxes are given)
before and after \textbf{Step-I} training. 
As illustrated in Tab.~\ref{tab:alignment}, 
the original plain CLIP model has certainly exhibited attribute classification on an elementary level. 
We can see a substantial improvement by further training on COCO-base and VAW-base, 
for example, from 46.15 to 57.39 AP for novel attribute classification, and 41.13 to 45.82 AP for novel category classification. 
Furthermore, by incorporating image-caption datasets in \textbf{Step-II} training, the performance has been improved to 69.03/65.17 AP on all attributes/categories. 
In the following experiments, we employ the model $\Phi_{\text{CLIP-Attr}}$ that exploits the COCO-Cap-Sub for Step-II training.

\vspace{4pt}
\noindent \textbf{Effect of Different Losses for Step-II Training.}
We investigate the performance variance by adjusting the different supervisions in Step-II training~(Eq.~\ref{eq:teacher_alignment2_loss}).
As illustrated in Tab.~\ref{tab:alignment_loss}, performance tends to grow monotonically with the increased supervision terms, from 66.92/55.21 to 69.03/65.17 AP on all attributes/categories, indicating that all supervision signals count.

\vspace{4pt}
\noindent \textbf{Knowledge Distillation.}
We validate the necessity for knowledge distillation while training OvarNet. Specifically, we experiment by training the model with federated loss only, and two other knowledge distillation approaches, \textit{i.e.}, the regional visual features (Feat.) from the visual encoder of $\Phi_{\text{CLIP-Attr}}$, like ViLD \cite{gu2021open}, and the prediction probability (Prob.) over all attributes from the matching scores of $\Phi_{\text{CLIP-Attr}}$. We achieve the distillation by constraining the regional visual feature or prediction probability of OvarNet to be the same as that of $\Phi_{\text{CLIP-Attr}}$, employing L2/L1 loss on features, 
and KL loss on probability. 
As shown in Tab.~\ref{tab:dist}, we make two observations, 
{\em first}, knowledge distillation is essential; {\em second}, knowledge gained from attribute prediction probabilities is more beneficial to improving performance on novel sets, \textit{e.g.}, from 51.87/33.17 to 56.43/54.10 AP when compared to L2 loss on visual features,
in particular for semantic classification on COCO.

\vspace{4pt}
\noindent \textbf{Different Architectures in $\Phi_{\text{CLIP-Attr}}$.}
We have evaluated the performance on different pre-trained CLIP architectures for attribute classification, such as R50 and ViT-B/16, and then conducted knowledge distillation from the different $\Phi_{\text{CLIP-Attr}}$ models.
As seen in Tab.~\ref{tab:arch}, 
both architectures perform competitively, with transformer-based architectures consistently outperforming the ConvNet ones.

\vspace{4pt}
\noindent \textbf{Updating OvarNet's Visual Backbone.}
We experiment by updating or freezing the OvarNet's visual backbone from different initializations at training. As shown in Tab.~\ref{tab:init}, initialising the visual backbone from aligned $\Phi_{\text{CLIP-Attr}}$ is advantageous, whereas finetuning or freezing it makes little difference. For efficiency, we opt to freeze the visual backbone in other experiments.

\hiddensubsection{Comparison with the State-of-the-Art}

\noindent \textbf{Benchmark on COCO and VAW.}
In Tab.~\ref{tab:sota}, we compare OvarNet to other attribute prediction methods and open-vocabulary object detectors on the VAW test set and COCO validation set. 
As there is no open-vocabulary attribute prediction method developed on the VAW dataset, 
we re-train two models on the {\em full} VAW dataset as an oracle comparison, namely, SCoNE~\cite{pham2021learning} and TAP~\cite{pham2022improving}. 
Our best model achieves 68.52/67.62 AP across all attribute classes for the box-given and box-free settings respectively.
On COCO open-vocabulary object detection, 
we compare with OVR-RCNN~\cite{zareian2021open}, ViLD~\cite{gu2021open}, Region CLIP~\cite{zhong2022regionclip}, PromptDet~\cite{feng2022promptdet}, and Detic~\cite{zhou2022detecting}, our best model obtains 54.10/35.17 AP for novel categories, surpassing the recent state-of-the-art ViLD-ens~\cite{gu2021open} and Detic~\cite{zhou2022detecting} by a large margin,
showing that attributes understanding is beneficial for open-vocabulary object recognition. Fig.~\ref{fig:results} shows some prediction results of OvarNet.

\begin{table}[t]
  \setlength{\tabcolsep}{2.5pt}
  \centering
  \resizebox{\linewidth}{!}{
  \footnotesize
  \begin{tabular}{l | c | c | *{3}{c}}
    \toprule
    \textbf{Method} & \textbf{Box Setting} & $\textbf{AP}_\textbf{all}$ & $\textbf{AP}_\textbf{head}$ & $\textbf{AP}_\textbf{medium}$ &  $\textbf{AP}_\textbf{tail}$ \\
    \midrule
    CLIP RN50 \cite{radford2021learning} & given & 15.8 & 42.5 & 17.5 & 4.2 \\
    CLIP VIT-B16 \cite{radford2021learning} & given &16.6 & 43.9 & 18.6 & 4.4 \\
    Open CLIP RN50 \cite{openclip} & given &11.8 & 41.0 & 11.7 & 1.4\\
    Open CLIP ViT-B16 \cite{openclip} & given &16.0 &45.4 &17.4 &3.8\\
    Open CLIP ViT-B32 \cite{openclip} & given &17.0 & 44.3 & 18.4 & 5.5\\
    ALBEF \cite{li2021align} & given & 21.0 & 44.2 & 23.9 & 9.4 \\
    BLIP \cite{li2022blip} & given & 24.3 & 51.0 & 28.5 & 9.7
    \\
    X-VLM \cite{zeng2021multi} & given & 28.1 & 49.7 & 34.2 & \cellcolor{gray!30}\textbf{12.9}
    \\
    OVAD \cite{bravo2022open} & given & 21.4 & 48.0 & 26.9 & 5.2
    \\
    CLIP-Attr RN50~(ours) & given & 24.1 & 54.8 &29.3 &6.7 
    \\
    CLIP-Attr ViT-B16~(ours) & given & 26.1 & 55.0  & 31.9  & 8.5
    \\
    OvarNet ViT-B16~(ours) & given & \cellcolor{gray!30}\textbf{28.6} & \cellcolor{gray!30}\textbf{58.6} & \cellcolor{gray!30}\textbf{35.5} & 9.5 
    \\\midrule
    OV-Faster-RCNN \cite{bravo2022open} & free & 14.1 & 32.6 & 18.3 & 2.5 \\
    Detic \cite{zhou2022detecting}& free & 13.3 & 44.4 & 13.4 & 2.3 \\
    OVD \cite{rasheed2022bridging}& free & 14.6 &33.5 &18.7 &2.8 \\
    LocOv \cite{bravo2022localized}& free & 14.9 & 42.8 & 17.2 & 2.2 \\
    OVR \cite{zareian2021open} & free & 15.1 & 46.3 & 16.7 & 2.1 \\
    OVAD \cite{bravo2022open} & free &  18.8& 47.7& 22.0& 4.6 \\
    OvarNet ViT-B16~(ours) & free & \cellcolor{gray!30}\textbf{27.2} & \cellcolor{gray!30}\textbf{56.8}  & \cellcolor{gray!30}\textbf{33.6}  & \cellcolor{gray!30}\textbf{8.9}\\
    \bottomrule
  \end{tabular}
  }
  \vspace{-8pt}
  \caption{Cross-dataset transfer evaluation on OVAD benchmark across all, head, medium, and tail attributes. Numbers are copied from~\cite{bravo2022open}.} 
  \vspace{-0.2cm}
  \label{tab:ovad}
\end{table}

\vspace{4pt}
\noindent \textbf{Cross-dataset Transfer on OVAD Benchmark.}
We compare with other state-of-the-art methods on OVAD benchmark~\cite{bravo2022open},
following the same evaluation protocol, we conduct zero-shot cross-dataset transfer evaluation with CLIP-Attr and OvarNet trained on COCO Caption dataset. 
Metric is average precision (AP) over different attribute frequency distributions, `head', `medium', and `tail'. 
As shown in Tab.~\ref{tab:ovad}, our proposed models largely outperform other competitors by a noticeable margin.

\vspace{4pt}
\noindent \textbf{Evaluation on LSA Benchmark.}
We evaluate the proposed OvarNet on the same benchmark proposed by Pham \textit{et al.}~\cite{pham2022improving}. 
As OpenTAP employs a Transformer-based architecture with object category and object bounding box as the additional prior inputs, we have evaluated two settings. One is the original OvarNet without any additional input information; the other integrates the object category embedding as an extra token into the transformer encoder layer in Sec.~\ref{sec:student_rpn}. 
As shown in Tab.~\ref{tab:lsa_benchmark}, OvarNet outperforms prompt-based CLIP by a large margin and surpasses OpenTAP~(proposed in the benchmark paper) under the same scenario, 
{\em i.e.}, with additional category embedding introduced. `Attribute prompt' means the prompt designed with formats similar to ``A photo of something that is [\underline{attribute}]", while `object-attribute prompt' denotes ``A photo of [\underline{category}] [\underline{attribute}]". For the `combined prompt', the outputs of the `attribute prompt' and the `object-attribute prompt' are weighted average.

\begin{table}[t]
  \setlength{\tabcolsep}{0.8pt}
  \centering
  \resizebox{\linewidth}{!}{
  \footnotesize
  \begin{tabular}{l c| *{3}{c} |  *{3}{c}}
    \toprule
    \multicolumn{1}{c}{\multirow{2}{*}{\textbf{Method}}} & \multicolumn{1}{c|}{\multirow{2}{*}{\textbf{Setting}}}& \multicolumn{3}{c|}{\multirow{1}{*}{\textbf{LSA common}}} & \multicolumn{3}{c}{\multirow{1}{*}{\textbf{LSA common $\rightarrow$ rare}}}\\
    \multicolumn{1}{c}{} & \multicolumn{1}{c|}{}& $\textbf{AP}_\textbf{base}$ & $\textbf{AP}_\textbf{novel}$ & $\textbf{AP}_\textbf{all}$ & 
    $\textbf{AP}_\textbf{base}$ & $\textbf{AP}_\textbf{novel}$ & $\textbf{AP}_\textbf{all}$ \\
    \midrule
    CLIP & attribute prompt & 2.53 & 3.37 &2.64 & 2.62 & 2.52 & 2.58\\
    CLIP & object-attribute prompt &0.97 & 1.56 & 1.04 & 1.16 & 0.73 & 0.97\\
    CLIP & combined prompt & 2.81 & 3.67 & 2.92 & 3.12 & 2.63 & 2.91\\
    OpenTAP & w/category prior & 14.34 & 7.62 & 13.59 & 15.39 & 5.37 & 10.91\\
    \midrule
    OvarNet & wo/category prior & 9.15 & 4.69 & 8.52 & 9.46 & 3.40 & 6.17 \\
    OvarNet & w/category prior & \cellcolor{gray!30}\textbf{15.57} & \cellcolor{gray!30}\textbf{8.05} & \cellcolor{gray!30}\textbf{14.84} & \cellcolor{gray!30}\textbf{16.74} & \cellcolor{gray!30}\textbf{5.48} & \cellcolor{gray!30}\textbf{11.83} \\
    \bottomrule
  \end{tabular}
  }
  \vspace{-6pt}
  \caption{Evaluation of LSA common and LSA common $\rightarrow$ rare. 
  Following the evaluation protocol in original paper~\cite{pham2022improving}, 
  all results are evaluated in a \textbf{box-given setting}.}\label{tab:lsa_benchmark}
  \vspace{-10pt}
\end{table}

\hiddensection{Conclusion}

In the paper, we consider the problem of open-vocabulary object detection and attribute recognition, {\em i.e.}, simultaneously localising objects and inferring their semantic categories and visual attributes. We start with a na\"ive two-stage framework (CLIP-Attr) that uses a pre-trained CLIP to classify the object proposals, 
to better align the object-centric visual feature with attribute concepts, we use learnable prompt vectors on the textual encoder side.
On the training side, we adopt a federated training strategy to exploit both object detection and attribute prediction datasets, and explore a weakly supervised training regime with external image-text pairs to increase the robustness for recognising novel attributes. Finally, for computational efficiency, we distill the knowledge of CLIP-Attr into a Faster-RCNN type model~(termed as OvarNet), while evaluating on four different benchmarks, {\em e.g.}, VAW, MS-COCO, LSA, and OVAD, we show that jointly training object detection and attributes prediction is beneficial for visual scene understanding, largely outperforming the existing approaches that treat the two tasks independently, demonstrating strong generalization ability to novel attributes and categories.

{\small
\bibliographystyle{ieee_fullname}
\bibliography{egbib}
}

\clearpage
\onecolumn
\appendix

\hypersetup{
colorlinks=true,
linkcolor=black
}
\renewcommand*\contentsname{Supplementary}
\tableofcontents
\newpage

\section{Base / Novel Attribute Set in VAW}

VAW \cite{pham2021learning} contains a large vocabulary of 620 attributes. In our experiments, considering that VAW attribute vocabulary has certain noise and semantic overlap, instead of taking all `tail’ attributes as the novel set, we sample half of the `tail’ attributes and 15\% of the `medium’ attributes as the novel set ($\mathcal{A}_{\text{novel}}$, 79 attributes) and the remaining as the base ($\mathcal{A}_{\text{base}}$, 541 attributes). The novel  attributes are: pulled back, smirking, muscular, holed, off white, littered, pepperoni, taupe, tucked in, bell shaped, multicolored, bronze, boiled, caucasian, silk, active, stormy, new, sprinkled, covered in sugar, side view, carried, overgrown, black metal, thatched, dotted, horned, shoeless, stucco, well dressed, barred, half filled, domed, vintage, hiding, gold framed, baked, reddish, rust colored, frizzy, nylon, scruffy, taking photo, opaque, violet, busy, foamy, relaxing, cubed, leaping, moss covered, chocolate, plastic, spreading arms, wispy, arch shaped, bent, bright green, black lettered, patchy, balancing, crocheted, furry, maroon, flat screen, classical, cloudless, partially visible, wearing scarf, orange, slender, eating, doorless, closed, shining, spotted, reflective, barren, wrapped.

\section{Summary of Dataset Statistics}

In our experiments, we take the standard MS-COCO 2017 \cite{lin2014microsoft} and VAW \cite{pham2021learning} for federated training, with the former for object category classification, and the latter for object attributes classification. In addition, we have harvested external image caption pairs on the COCO and VAW dictionaries from the CC 3M \cite{sharma2018conceptual} and COCO Captions \cite{chen2015microsoft} for training CLIP-Attr. As for evaluation, we also include two additional benchmarks~(LSA~\cite{pham2022improving} and OVAD~\cite{bravo2022open}) using official settings in their papers. Tab.~\ref{tab:dataset} contains the detailed statistics for all relevant datasets.

\begin{table}[htb]
  \centering
  \small
  \resizebox{1\linewidth}{!}{
  \begin{tabular}{l *{5}{c}}
    \toprule
    Dataset & Train & Eval. & Description & Images & Categories/Attributes\\
    \midrule
    MS-COCO & - & - & original COCO detection dataset \cite{lin2014microsoft} & 118K & 80 \\
    VAW & - & - & original VAW attribute prediction dataset \cite{pham2021learning} & 58K & 620 \\
    COCO Cap & - & - & COCO Caption dataset \cite{chen2015microsoft} & 118K & image-text pairs \\
    CC 3M & - & - & Conceptual Captions 3M dataset \cite{sharma2018conceptual} & 3M & image-text pairs \\
    LSA & \checkmark & \checkmark & original LSA dataset for training and evaluating \cite{pham2022improving} & 420K & 5526 \\
    OVAD & \ding{55} & \checkmark & original OVAD benchmark for evaluating \cite{bravo2022open} & 2K & 117 \\
    \midrule
    COCO-base & \checkmark & \ding{55} & base categories on COCO dataset \cite{bansal2018zero} & 107K & 48 \\
    VAW-base & \checkmark & \ding{55} & base attributes on VAW dataset & 58K & 541 \\
    CC-3M-sub & \checkmark & \ding{55} &  available online pairs filtered by the dictionaries & 1M & noise \\
    COCO-Cap-sub & \checkmark & \ding{55} & image-text pairs filtered by the dictionaries & 118K & noise \\
    \midrule
    COCO-novel & \ding{55} & \checkmark & 65 categories on COCO val dataset setted by \cite{bansal2018zero} & 5K & 65 \\
    VAW-novel & \ding{55} & \checkmark & all attributes in VAW test dataset & 10K & 620 \\
    \bottomrule
  \end{tabular}
  }
  \vspace{-3pt}
  \caption{A summary of dataset statistics} \label{tab:dataset}
  \vspace{-0.2cm}
\end{table}

\newpage 
\section{Ablation Study}
In this section, we provide additional ablation studies that are not included in the main paper, due to space limitations.

\vspace{5pt}
\noindent \textbf{Effect of Prompt Vectors.}
We have conducted experiments by varying numbers of prompt vectors in the CLIP-Attr,
all results are obtained from the model after Step-I training.
Prompt vectors are split evenly and placed before, between, and after the attribute and parent-class attribute word. As illustrated in Tab.~\ref{tab:num_prompt}, 
our model is relatively robust to the different number of prompt vectors.

\begin{table}[htbp]
\centering
\small
\resizebox{0.4\linewidth}{!}{
  \begin{tabular}{c | *{2}{c} | *{2}{c}}
    \toprule
    \multicolumn{1}{c|}{\multirow{2}{*}{\textbf{\# prompts}}} & \multicolumn{2}{c|}{\multirow{1}{*}{\textbf{VAW}}} & \multicolumn{2}{c}{\multirow{1}{*}{\textbf{COCO}}}\\
    \multicolumn{1}{c|}{} & $\textbf{AP}_\textbf{novel}$ & $\textbf{AP}_\textbf{all}$ & $\textbf{AP}_\textbf{novel}$ & $\textbf{AP}_\textbf{all}$\\
    \midrule
    3 & 56.94 & 66.39 & 45.27 & 54.45\\
    9 & 57.13 & 66.72 & 45.50 & 54.86\\
    15 & 57.24 & 66.80 & 45.77 & 55.05\\
    30 & 57.39 & 66.92 & \cellcolor{gray!30}45.82 & 55.21\\
    60 & \cellcolor{gray!30}57.41 & \cellcolor{gray!30}67.11 & 45.79 & \cellcolor{gray!30}55.32\\
    \bottomrule
  \end{tabular}
  }
  \vspace{-3pt}
  \caption{Effect of different numbers of prompt vectors in CLIP-Attr with first step alignment.}
  \vspace{-0.1cm}
\label{tab:num_prompt}
\end{table}

\noindent \textbf{Effect of Different Pooling Strategies.}
We adopt different architectures to extract regional visual features, including CNN and Transformer. 
The CNN architecture contains three convolution blocks with a stride of 2, followed by average pooling and a 2-layer MLP, while attentional pooling consists of a 4-layer transformer encoder. 
As illustrated in Tab.~\ref{tab:att}, 
we observe that employing the Transformer with attentional pooling to extract regional visual representation significantly outperforms the convolutional blocks w/ or w/o knowledge distillation.

\begin{table}[htbp]
  \centering
  \resizebox{0.57\linewidth}{!}{
  \begin{tabular}{l c | *{2}{c} | *{2}{c}}
    \toprule
    \multicolumn{1}{c}{\multirow{2}{*}{\textbf{Visual head}}} &
    \multicolumn{1}{c|}{\multirow{2}{*}{\textbf{Distil.}}} &\multicolumn{2}{c|}{\multirow{1}{*}{\textbf{VAW}}} & \multicolumn{2}{c}{\multirow{1}{*}{\textbf{COCO}}}\\
    \multicolumn{1}{c}{} & \multicolumn{1}{c|}{} & $\textbf{AP}_\textbf{novel}$ & $\textbf{AP}_\textbf{all}$ & $\textbf{AP}_\textbf{novel}$ & $\textbf{AP}_\textbf{all}$\\
    \midrule 
    CNN Blocks-AvgPool & none & 48.69 & 60.58 & 28.63 & 58.46 \\
    Transformer-AttnPool & none & 50.53 & 61.74 & 30.43 & 59.83 \\
    CNN Blocks-AvgPool & Prob. KL & 53.35 & 65.80 & 49.40 & 62.15 \\
    Transformer-AttnPool & Prob. KL & \cellcolor{gray!30}56.43 & \cellcolor{gray!30}68.52 & \cellcolor{gray!30}54.10 & \cellcolor{gray!30}67.23 \\
    \bottomrule
  \end{tabular}
  }
  \vspace{-3pt}
  \caption{Ablation study on different pooling strategy with a box-given setting.} \label{tab:att}
    \vspace{-0.1cm}
\end{table}

\noindent \textbf{Effect of Transformer Encoder Layers.}
Here, we also perform an ablation investigation on different numbers of transformer encoder layers in attentional pooling using probability distillation. As indicated in Tab. \ref{tab:trans_layer}, the number of transformer encoder layers has only a slight influence on performance, and a 4-layer transformer is sufficient to  achieve comparable performance.

\begin{table}[!htb]
  \centering
  \small
  \resizebox{0.38\linewidth}{!}{
  \begin{tabular}{c | *{2}{c} | *{2}{c}}
    \toprule
    \multicolumn{1}{c|}{\multirow{2}{*}{\textbf{\# layers}}} &\multicolumn{2}{c|}{\multirow{1}{*}{\textbf{VAW}}} & \multicolumn{2}{c}{\multirow{1}{*}{\textbf{COCO}}}\\
    \multicolumn{1}{c|}{} & $\textbf{AP}_\textbf{novel}$ & $\textbf{AP}_\textbf{all}$ & $\textbf{AP}_\textbf{novel}$ & $\textbf{AP}_\textbf{all}$\\
    \midrule
    2 & 55.02 & 67.19 & 52.28 & 65.33 \\
    4 & 56.43 & \cellcolor{gray!30}68.52 & \cellcolor{gray!30}54.10 & \cellcolor{gray!30}67.23
    \\
    6 & \cellcolor{gray!30}56.71 & 68.26 & 53.90 & 67.17 \\
    \bottomrule
  \end{tabular}
  }
  \vspace{-3pt}
  \caption{Different number of transformer encoder layers in attentional pooling under a box-given setting.} \label{tab:trans_layer}
  \vspace{-0.1cm}
\end{table}

\newpage
\section{Qualitative Results}

\begin{figure}[htbp]
\centering
\includegraphics[width=0.98\linewidth]{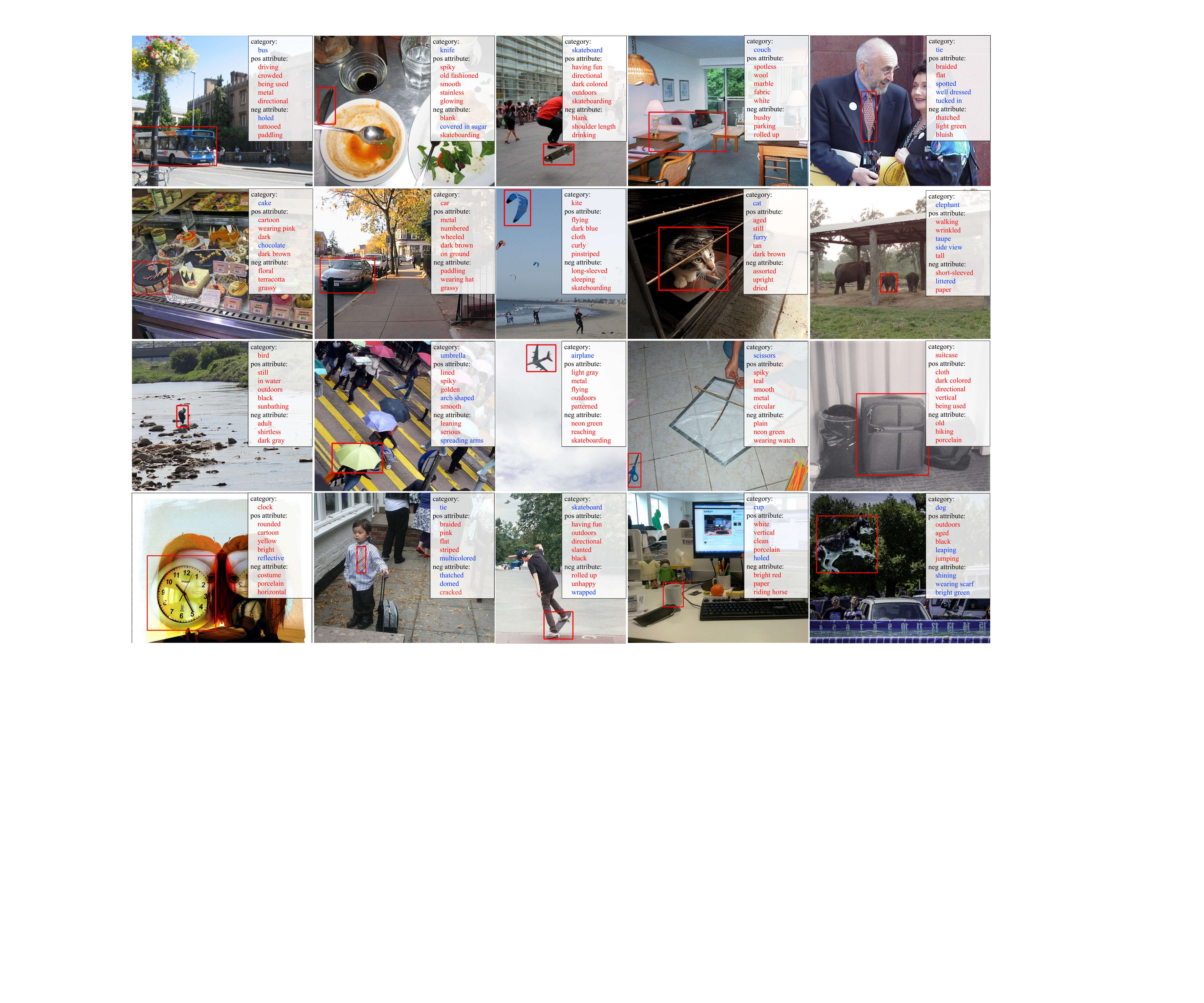}
\caption{Visualization of prediction results. \textcolor[rgb]{1,0,0}{\textbf{Red}} denotes the base category/attribute \textit{i.e.},  seen in the training set, while  \textcolor[rgb]{0.07,0.19,0.96}{\textbf{blue}} represents the novel category/attribute unseen in the training set. The first two rows are samples from the VAW test set, while the last two rows are from the COCO val set.} \label{fig:vis_results}
\end{figure}

In Fig.~\ref{fig:vis_results}, we show the qualitative results of OvarNet on VAW and MS-COCO benchmarks. OvarNet is capable of accurately localizing, recognizing, and characterizing objects based on a broad variety of novel categories and attributes.

\clearpage 
\section{Failure Cases \& Limitations}

\begin{figure}[!htb]
\centering
\includegraphics[width=0.98\linewidth]{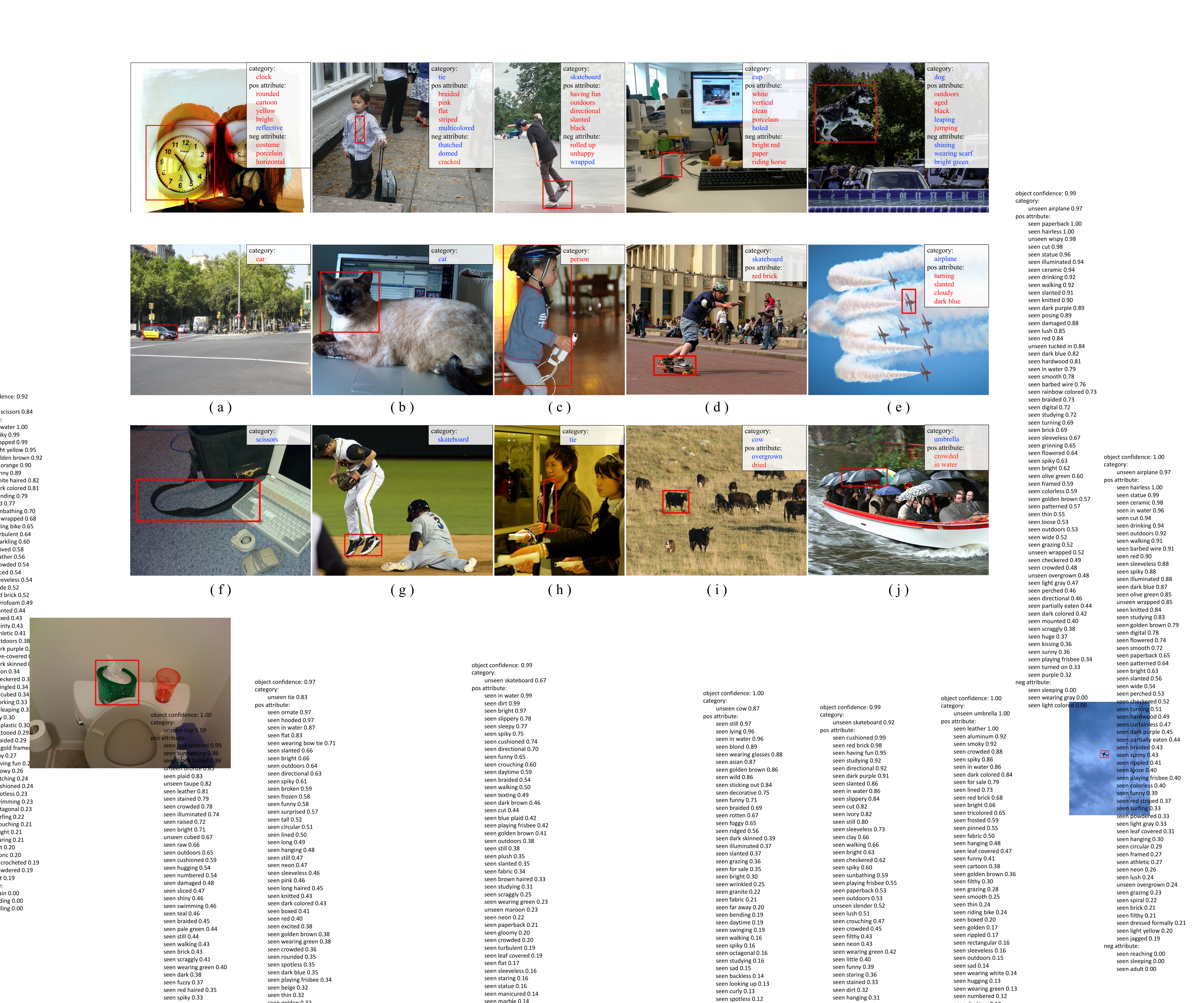}
\vspace{-0.4cm}
\caption{Visualization of failure cases. \textcolor[rgb]{1,0,0}{\textbf{Red}} denotes the base category/attribute \textit{i.e.},  seen in the training set, while  \textcolor[rgb]{0.07,0.19,0.96}{\textbf{blue}} represents the novel category/attribute unseen in the training set.} \label{fig:fail}
\end{figure}

In this section, 
we present some analysis of failure cases, as depicted in Fig.~\ref{fig:fail}, hoping it will inspire future works.
Generally speaking, we observe three major failure types: 
partial localisation, \textit{e.g.}, (a), (b), and (c); 
misclassification for the semantic category, \textit{e.g.}, (f), (g), and (h); 
partially inaccurate attribute descriptions, \textit{e.g.}, (d), (e), (i), and (j).

\vspace{3pt}
\noindent \textbf{Partial localisation} refers to the cases with inaccurate localisation,
as shown in Fig.~\ref{fig:fail} (a), (b), and (c). We discover that a target may be represented by many bounding boxes and that some bounding boxes only encompass a portion of the object, yet they are not removed after non-maximum suppression and have high confidence in the classification score. We believe that partial localisation is mostly caused by the localisation component, and category classification is achieved by following the guidance of response from the partial area in the object.

\vspace{3pt}
\noindent \textbf{Misclassification of semantic category} denotes that an object is recognized with low confidence, as illustrated in Fig.~\ref{fig:fail} (f), (g), and (h). Given a box proposal, it is difficult to remove the none-object error boxes, 
as the classifier may also be able to infer the category with context information. For example, Fig.~\ref{fig:fail} (h) shows a failure case of a tie. 

\vspace{3pt}
\noindent \textbf{Partially inaccurate attribute description} denotes inaccurate attribute prediction, as illustrated in Fig.~\ref{fig:fail} (d), (e), (i), and (j). We find the model appears to assign representations of the surrounding environment or background to the object in some circumstances.

\clearpage

\end{document}